\documentclass[dvipsnames]{bmvc2k}

\usepackage{makecell}
\usepackage{pifont}
\usepackage{enumerate}
\usepackage{amsmath, amssymb, amsfonts}
\usepackage{algorithm}
\usepackage{algorithmicx, algpseudocode}
\usepackage{booktabs}
\usepackage{multirow} % 确保加载了这个包
\usepackage{array}
\usepackage{float}
\usepackage{graphicx}
\usepackage{threeparttable}
\usepackage{wrapfig}

\title{RPD-Diff: Region-Adaptive Physics-Guided Diffusion Model for Visibility Enhancement under Dense and Non-Uniform Haze}

\addauthor{Ruicheng Zhang$^*$}{}{1,2}
\addauthor{Puxin Yan$^*$}{}{1,2}
\addauthor{Zeyu Zhang}{}{3}
\addauthor{Yicheng Chang}{}{1,2}
\addauthor{Hongyi Chen}{}{1,2}
\addauthor{Zhi Jin$^\dagger$}{}{1,2}

\addinstitution{
 School of Intelligent Systems Engineering, Shenzhen Campus of Sun Yat-sen University,
 Shenzhen, Guangdong 518107, China
}
\addinstitution{
 Guangdong Provincial Key Laboratory of Fire Science and Intelligent Emergency Technology, 
 Shenzhen 518107, China.
}
\addinstitution{
 The Australian National University,\\
 Canberra, Australia
}

\runninghead{Zhang, Yan, et al.}{RPD-Diff: Region-Adaptive Physics-Guided Diffusion}

\begin{document}

\maketitle

% --- Add this block for all footnotes on the first page ---
\begingroup
\renewcommand\thefootnote{}\footnotetext{
    \hspace{-2em} % Negative space to align with the margin
    $^*$Equal contribution. $^\dagger$Corresponding author. \\
    This work was supported by Guangdong S\&T Program No. 2024B1111060003, the National Natural Science Foundation of China under Grant U24A20251, 62071500, and Shenzhen Science and Technology Program under Grant JCYJ20230807111107015.
}
\endgroup
% --- End of footnote block ---

\vspace{-2em}
\begin{abstract}
Single-image dehazing under dense and non-uniform haze conditions remains challenging due to severe information degradation and spatial heterogeneity. Traditional diffusion-based dehazing methods struggle with insufficient generation conditioning and lack of adaptability to spatially varying haze distributions, which leads to suboptimal restoration. To address these limitations, we propose RPD-Diff, a Region-adaptive Physics-guided Dehazing Diffusion Model for robust visibility enhancement in complex haze scenarios. RPD-Diff introduces a Physics-guided Intermediate State Targeting (PIST) strategy, which leverages physical priors to reformulate the diffusion Markov chain by generation target transitions, mitigating the issue of insufficient conditioning in dense haze scenarios. Additionally, the Haze-Aware Denoising Timestep Predictor (HADTP) dynamically adjusts patch-specific denoising timesteps employing a transmission map cross-attention mechanism, adeptly managing non-uniform haze distributions. Extensive experiments across four real-world datasets demonstrate that RPD-Diff achieves state-of-the-art performance in challenging dense and non-uniform haze scenarios, delivering high-quality, haze-free images with superior detail clarity and color fidelity.
\end{abstract}

%-------------------------------------------------------------------------
\vspace{-2em}
\section{INTRODUCTION}
\label{sec:intro}
Single-image dehazing is crucial for downstream visual tasks \cite{DC-Scence} such as object detection, autonomous driving, and surveillance \cite{FDG-Diff}, thus attracting extensive research interest. However, existing dehazing methods \cite{FDG-Diff,FFA,FSDGN,2011Single,2016Robust,2017AOD,2017Fast,DAPM,dehazeformer,dehamer,MSBDN,jinzhi_ICCV}, although effective in synthetic datasets or under specific real-world conditions, could struggle with the challenges posed by dense and non-uniform haze frequently encountered in practice. Such challenging haze conditions are characterized by severe information degradation and significant spatial variations in physical properties. This combination of degraded content and color information, along with the non-uniform nature of the haze across the image, substantially complicates the restoration process.

\begin{wrapfigure}{r}{0.65\textwidth}
  \centering
  \vspace{-0.5em}
  \includegraphics[width=0.65\textwidth]{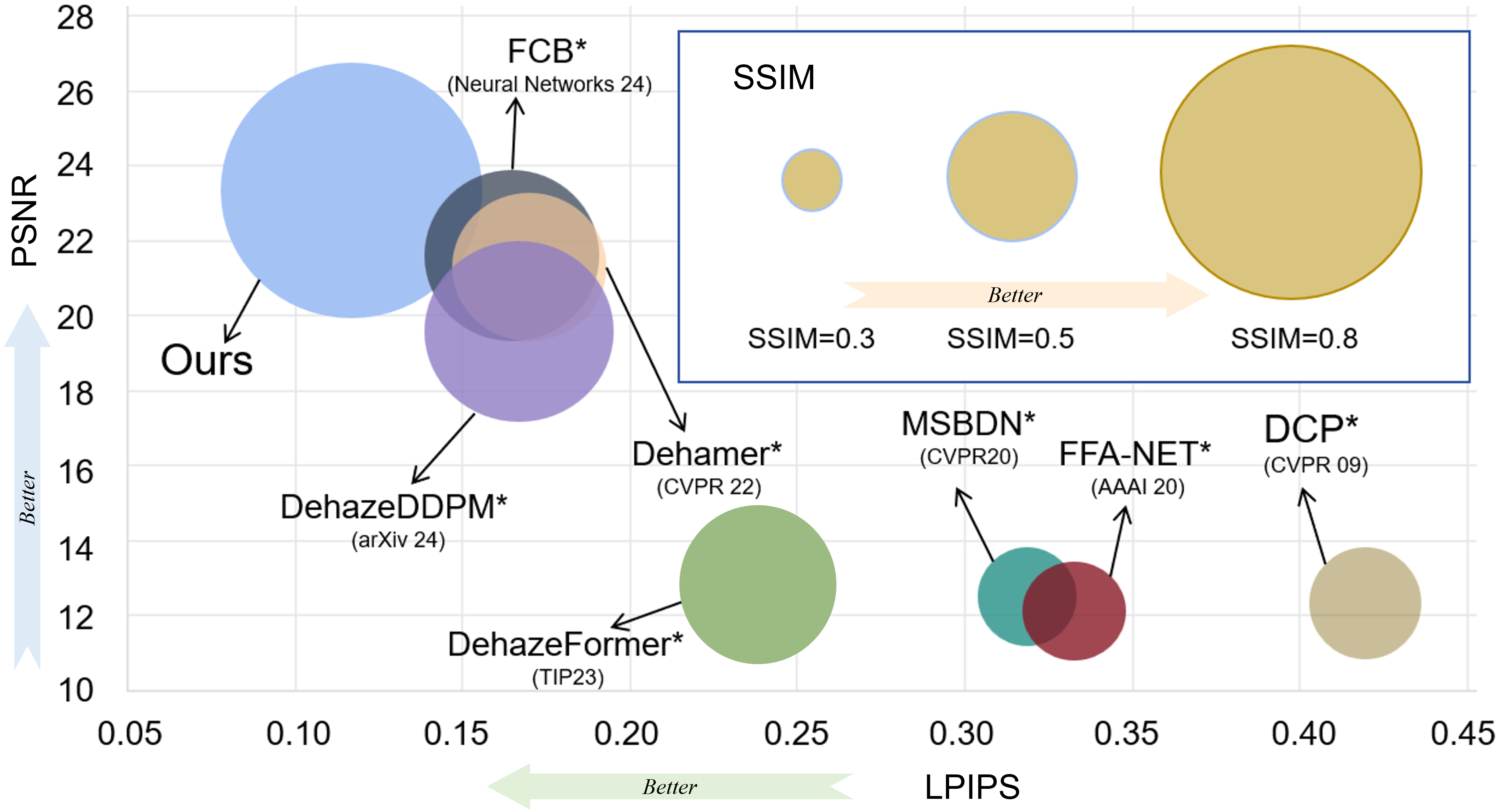}
  \vspace{-1em}
  \caption{Performance of our approach compared to state-of-the-art methods on the NH-Haze dataset \cite{nhaze}. RPD-Diff achieves state-of-the-art performance, outperforming previous methods on both distortion-based (PSNR$\uparrow$, SSIM$\uparrow$~\cite{psnrssim}) and perceptual (LPIPS$\downarrow$~\cite{LPIPS}) metrics.}
  \label{fig:performance}
  \vspace{-0.5em}
\end{wrapfigure}

Denoising Diffusion Probabilistic Models (DDPMs) \cite{DDPM} have demonstrated significant success in high-quality image generation and translation across various domains \cite{zhou2025fireedit,DriftRec,NEURIPS2021_49ad23d1}, shedding light on image dehazing.However, existing applications of DDPMs to the dehazing task \cite{WANG2024106281, yu2024highqualityimagedehazingdiffusion} exhibit critical limitations, particularly when confronted with dense and non-uniform haze:
(1) \textbf{\textit{Insufficient Conditioning}}: Current DDPM-based dehazing methods typically condition the diffusion process on the hazy input image to directly predict the clear counterpart. However, this strategy proves suboptimal in dense haze scenarios where significant information degradation severely compromises the hazy image's utility as reliable conditioning signals. Guided by such weak conditioning signals, the direct generation of a clear image from pure Gaussian noise becomes challenging, potentially leading to the loss of fine content details and slow training convergence.
(2) \textbf{\textit{Lack of Spatial Adaptivity}}: Standard DDPM frameworks generally lack explicit mechanisms to perceive and adapt to the spatial heterogeneity inherent in non-uniform haze, where different image regions exhibit varying haze density and associated restoration difficulties \cite{yu2024highqualityimagedehazingdiffusion}. These models typically apply uniform processing across the image, which hinders the accurate modeling and effective restoration required for complex, spatially varying haze distributions.

To address these limitations, we propose RPD-Diff, a Region-adaptive Physics-guided Dehazing Diffusion Model designed for visibility enhancement in dense and non-uniform haze conditions. The denoising Markov chain of DDPM progressively transforms a Gaussian noise distribution towards the target data distribution, with quality of the results directly influenced by the definition of the denoising target and the denoising step schedule. Recognizing that the physical characteristics of dense and non-uniform haze fundamentally alter the optimal way to define these targets and schedule these steps across an image, RPD-Diff introduces two key components designed to adapt the diffusion process accordingly:
(1) \textbf{\textit{Physics-guided Intermediate State Targeting (PIST)}} strategy mitigates the problem of insufficient conditioning associated with direct clear image generation, particularly in dense haze scenarios. This strategy adaptively adjusts the Markov chain transitions within the diffusion process for different image regions guided by the physical prior. The modified chain initially targets the reconstruction of hazy images and gradually transitions towards clear image generation, leveraging a haze-aware intermediate state function facilitating the shift. This transforms the challenging task of direct clear image generation into a more manageable reconstruction task of intermediate hazy state, particularly during initial denoise stages. By reframing the task from direct generation to intermediate reconstruction, PIST effectively improves the restoration quality with fewer denoising times, especially in dense haze scenarios.
(2) \textbf{\textit{Haze-Aware Denoising Timestep Predictor (HADTP)}} is proposed to handle the spatial physical heterogeneity inherent in non-uniform haze. HADTP employs a transmission map cross-attention mechanism to dynamically adjust the denoising timesteps and corresponding intensities of different patches based on local degradation degree. This enables the model to tailor the denoising process to localized haze characteristics, facilitating precise adaptation to non-uniform haze distributions.

Our main contributions are summarized as follows:
\begin{enumerate}
  \item We propose a Region-adaptive Physics-guided Dehazing Diffusion Model (RPD-Diff), which takes full advantage of the physical priors associated with dense and non-uniform haze to achieve high-quality visibility enhancement.
  \item We introduce a Physics-guided Intermediate State Targeting (PIST) strategy that adaptively refines region-specific Markov chain transitions guided by physical priors. This strategy simplifies image restoration via task transformation, effectively addressing the insufficient conditioning signal problem in dense haze scenarios.
  \item We develop the adaptive Haze-Aware Denoising Timestep Predictor (HADTP) that dynamically adjusts patch-specific denoising timestep leveraging a transmission map cross-attention mechanism, which effectively addresses the spatial physical heterogeneity inherent in non-uniform haze.
\end{enumerate}

Experimental results on the four public real-world datasets demonstrate that our proposed methods outperform other SOTA methods. In addition, our method shows better visual effects on images with dense haze and non-uniform scenes.

\begin{figure*}[tp] 
    \centering
    \vspace{-0.05cm}
    \setlength{\abovecaptionskip}{0.0cm}
    \setlength{\belowcaptionskip}{-0.1cm}
    \includegraphics[width=\textwidth]{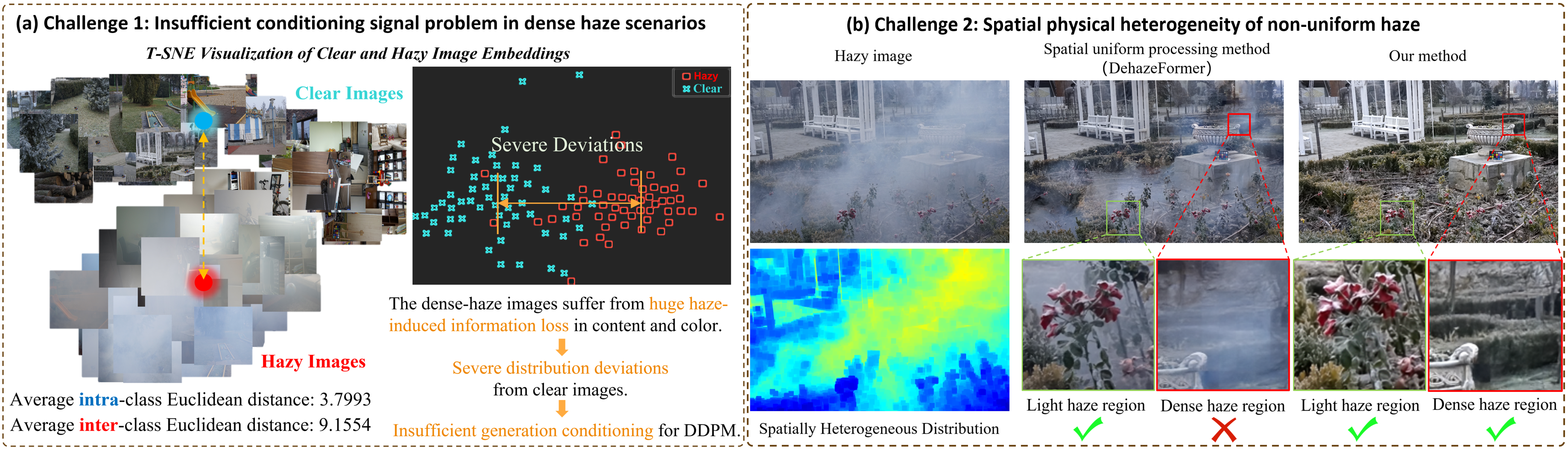}
    \vspace{-2.1em}
    \caption{Challenges in dehazing under scenarios of dense and non-uniform haze.}
    \label{fig:challenges}
    \vspace{-0.4cm}
\end{figure*}

\section{RELATED WORK}
Early dehazing methods often relied on the atmospheric scattering model (ASM) \cite{asm} and hand-crafted priors derived from statistical analyses of hazy and clear image pairs, such as the dark channel prior \cite{2011Single}, color line prior \cite{Raanan2014Dehazing}, sparse gradient prior \cite{2016Robust}, and maximum reflectance prior \cite{2017Fast}. These approaches attempt to estimate the transmission map and atmospheric light based on these priors to invert the ASM. However, these prior-based methods frequently exhibit limited robustness and generalization capabilities, primarily because their underlying assumptions may not hold true across diverse real-world scenarios, especially those involving dense or non-uniform haze.

Driven by the remarkable success of deep learning, methods employing neural networks have emerged as the dominant approach in image dehazing, achieving substantial advancements. These approaches, predominantly based on CNN or Transformer architectures, learn implicit image priors to aid in estimating ASM parameters \cite{Cai_Xu_Jia_Qing_Tao_2016, Liu_Pan_Ren_Su_2019} or directly model the end-to-end mapping from hazy to clear images \cite{dehamer, FSDGN}. Despite significant improvement in dehazing quality and robustness facilitated by sophisticated network designs and training strategies, limitations persist in handling dense and non-uniform haze. CNN-based methods \cite{FFA,MSBDN,FSDGN} can struggle with capturing long-range dependencies necessary for the restoration of severely degraded regions in non-uniform haze; while Transformer-based methods \cite{jinzhi_ICCV,dehazeformer,dehamer} excel at modeling global context, they can be susceptible to feature interference in dense haze scenarios, hindering accurate global information extraction and potentially causing color shifts or artifacts \cite{DAPM}.

More recently, Denoising Diffusion Probabilistic Models (DDPMs), recognized as powerful generative frameworks capable of high-fidelity image generation \cite{FDG-Diff}, have received significant attention in the dehazing field. For instance, DehazeDDPM \cite{yu2024highqualityimagedehazingdiffusion} introduces a conditional DDPM framework integrating physical modeling principles. GDP \cite{GDP} leveraged a pre-trained DDPM to perform high-fidelity unsupervised haze removal. Nevertheless, existing diffusion-based dehazing methods can still exhibit limitations when applied to challenging dense or non-uniform haze conditions. These include insufficient conditioning guidance derived from severely degraded hazy inputs and a lack of inherent adaptability to spatially varying haze density across the image. Such limitations can potentially lead to incomplete haze removal or the generation of visual artifacts \cite{DAPM}, motivating the need for more adaptive and robust diffusion-based approaches.

\begin{figure*}[tp] 
    \centering
    \setlength{\abovecaptionskip}{0.0cm}
    \setlength{\belowcaptionskip}{-0.1cm}
    \vspace{-0.05cm}
    \includegraphics[width=\textwidth]{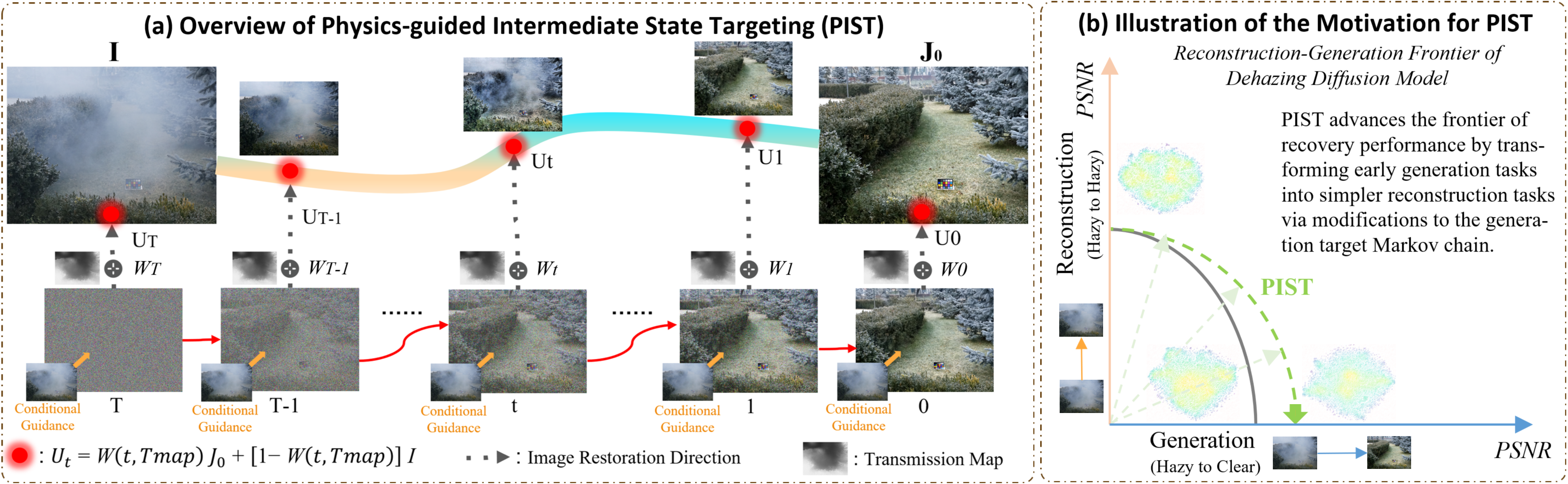}
    \vspace{-2.2em}
    \caption{Overview and motivation of Physics-guided Intermediate State Targeting (PIST).}
    \label{fig:pist}
    \vspace{-0.4cm}
\end{figure*}

\section{METHOD}
\subsection{Overview}
To address the challenges posed by dense and non-uniform haze in real-world images, which exhibit spatially varying degradation characteristics, RPD-Diff employs a patch-based diffusion framework \cite{patchDDPM}, which aligns well with the localized nature of complex haze variations and facilitates the seamless integration of our proposed components: Physics-guided Intermediate State Targeting (PIST) conducts patch-specific diffusion trajectory adaption guided by physical priors, mitigating the insufficient conditioning signal problem in dense haze scenarios; and Haze-Aware Denoising Timestep Predictor (HADTP) dynamically adjusts patch-specific denoising timesteps employing a transmission map cross-attention mechanism, effectively addressing the spatial physical heterogeneity of non-uniform haze.

\subsection{Patch-based Diffusive Dehazing Framework}
In real-world scenarios, atmospheric haze frequently manifests as a non-uniform distribution correlated with scene depth, which leads to spatially heterogeneous degradation levels and characteristics across the image. However, many conventional learning-based dehazing methods neglect this spatial variability, typically applying uniform processing globally \cite{DAPM}. Consequently, when applied to images affected by dense or non-uniform haze, these methods often produce localized over-enhancement or color distortions \cite{DAPM}.

\begin{wrapfigure}{r}{0.35\textwidth}
  \centering
  \vspace{-0.5em}
  \includegraphics[width=0.35\textwidth]{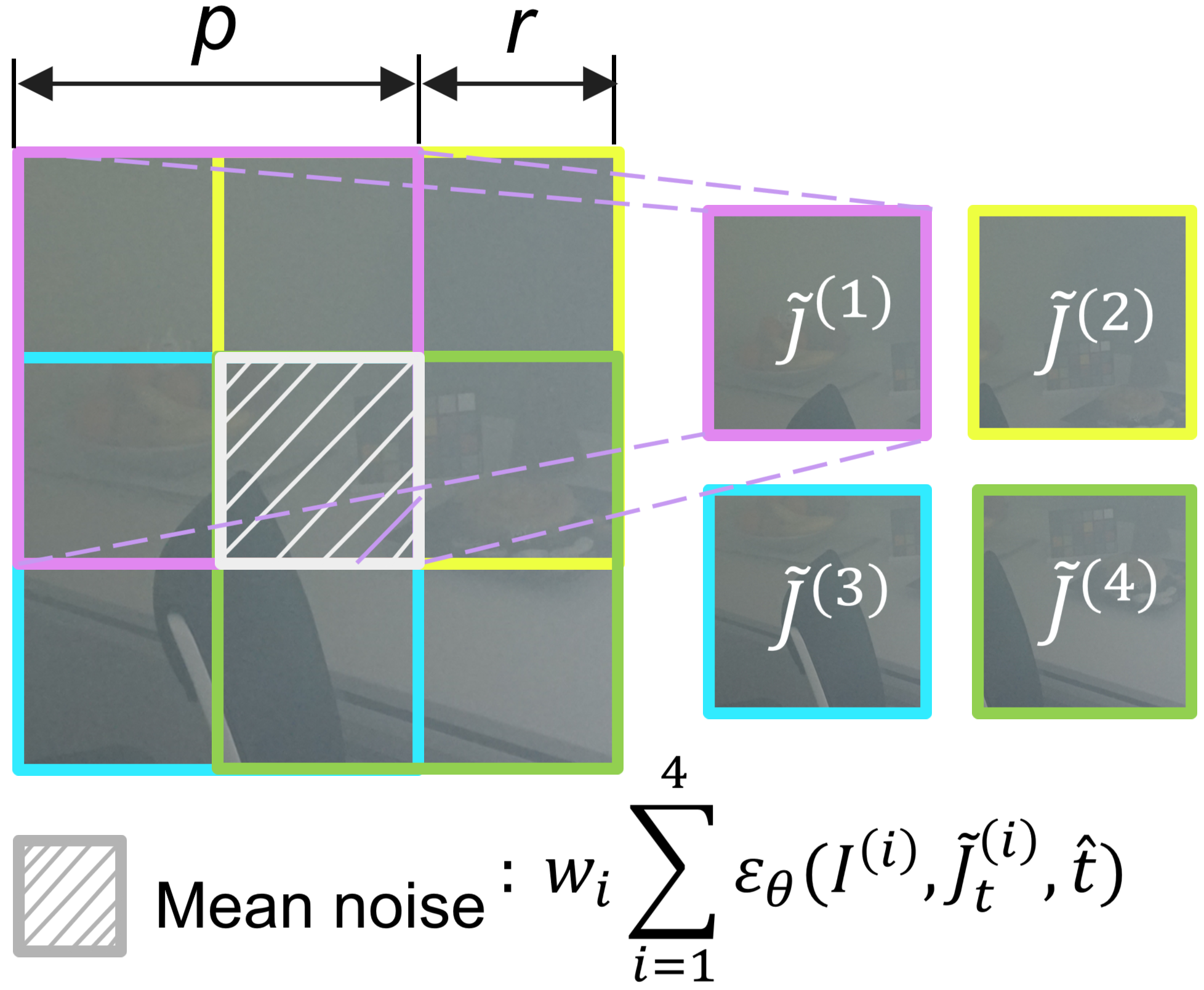}
  \vspace{-1em}
  \caption{Sampling for overlapping patches.}
  \label{fig:patch_base}
  \vspace{-0.5em}
\end{wrapfigure}

To effectively address such spatially heterogeneous degradations, RPD-Diff adopts a patch-based restoration framework, enabling localized dehazing tailored to regional characteristics. As illustrated in Fig.~\ref{fig:patch_base}, it extracts overlapping patches of size \( p \times p \) from the hazy input \( I \) with a stride \( r<p \), ensuring spatial continuity between adjacent patches. During the reverse diffusion process, the denoising network \( f_\theta \), incorporating the proposed PIST and HADTP mechanisms, estimates the noise \( \epsilon_\theta \) for each patch conditioned on local haze properties. For pixels within overlapping regions, we introduce a learnable parameter-based superposition mechanism to aggregate noise estimates from contributing patches:
\begin{equation}
\label{eq:learnable_noise}
\setlength\abovedisplayskip{3pt}
\setlength\belowdisplayskip{0pt}
\epsilon_\theta^{(i)} = f_\theta \left( J_t^{(i)}, \tilde{I}^{(i)}, t \right)_{\text{PIST}}^{\text{HADTP}}, \quad \overline{\epsilon}_\theta = \sum_{i=1}^{n} w_i \epsilon_\theta^{(i)},
\end{equation}
where \( J_t^{(i)} \) denotes the noisy patch at timestep \( t \), \( \tilde{I}^{(i)} \) represents the corresponding hazy patch condition, \( n \) is the number of overlapping patches covering a given pixel, and \( w_i \) are learnable weights satisfying \( \sum_{i=1}^{n} w_i = 1 \). This learnable superposition strategy offers several advantages. First, it mitigates discontinuities at patch boundaries by dynamically prioritizing noise estimates from patches with more reliable local information. Second, it enhances the model's ability to capture long-range dependencies by facilitating context sharing across adjacent patches, which is critical for coherent restoration of the entire image. Finally, the incorporation of learnable weights enables the framework to better handle complex haze patterns, improving robustness against artifacts compared to the conventional averaging approach. Notably, this patch-based framework not only addresses spatially heterogeneous degradation effectively but also offers significant computational advantages. Processing smaller patches individually is substantially more efficient than performing global diffusion on a full-resolution image~\cite{patchDDPM}, leading to a reduced computational footprint.

\subsection{Physics-guided Intermediate State Targeting}

\begin{figure*}[tp] 
    \centering
    \vspace{-0.05cm}
    \setlength{\abovecaptionskip}{0.0cm}
    \setlength{\belowcaptionskip}{-0.4cm}
    \includegraphics[width=\textwidth]{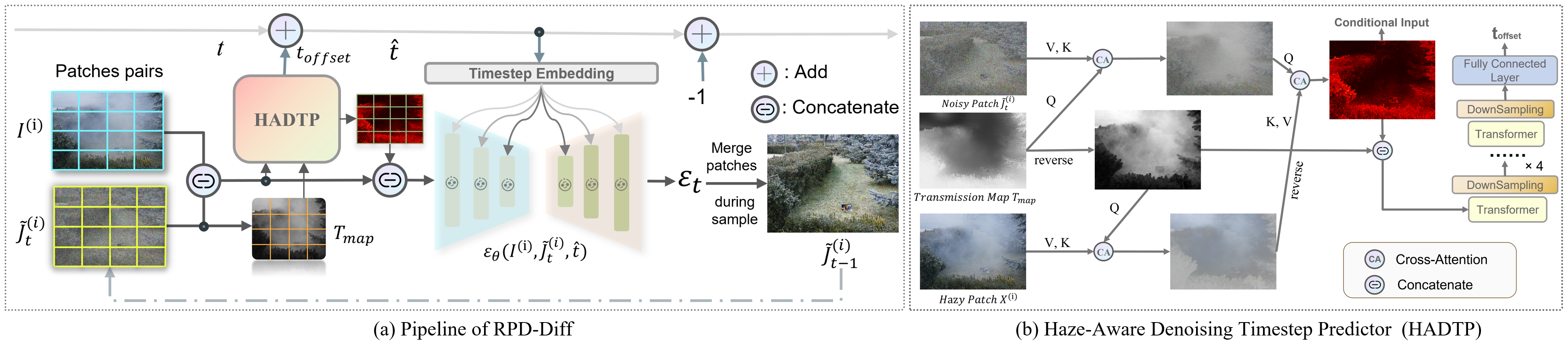}
    \vspace{-2.1em}
    \caption{The architecture of RPD-Diff. At each timestep $t$, the Haze-Aware Denoising Timestep Predictor (HADTP) first adjusts the denoising timestep and generates enhanced conditional input employing a transmission map cross-attention mechanism. Subsequently, the model restores the image \( J_{\hat{t}-1} \) from \( J_{\hat{t}} \), utilizing the estimated even noise $\bar{\epsilon}_{\hat{t}}$.}
    \label{fig:pipeline}
    \vspace{-0.2cm}
\end{figure*}

Standard Denoising Diffusion Probabilistic Models (DDPMs) for image dehazing typically aim to recover the clear image \( J_0 \) directly from a noisy state \( J_t \), conditioned on the hazy image \( I \). However, in scenarios with dense haze, severe information loss renders \( I \) an unreliable conditional input, often leading to suboptimal restoration quality and slow convergence. To overcome this limitation, we introduce the \textbf{P}hysics-guided \textbf{I}ntermediate \textbf{S}tate \textbf{T}argeting (PIST) strategy. As illustrated in Fig.~\ref{fig:pist}, PIST reformulates the diffusion process by dynamically adjusting the target state based on physical haze properties, thereby enhancing both restoration fidelity and efficiency. This approach improves performance by transforming the challenging initial generation steps into simpler reconstruction tasks, effectively altering the target of the generative Markov chain.

Specifically, PIST introduces a spatially and temporally adaptive intermediate target state \( U_t^{i}(x) \) for each image patch \( i \). This state interpolates between the clear image \( J_0^{i}(x) \) and the hazy image \( I^{i}(x) \), guided by the local haze density:
\begin{equation}
\label{eq:ut_definition_reorder}
U_t^{i}(x) = W^{i}(t, T^{i}_{\text{map}}(x)) \cdot J_0^{i}(x) + \big(1 - W^{i}(t, T^{i}_{\text{map}}(x))\big) \cdot I^{i}(x),
\end{equation}
where $x$ denotes the spatial coordinate within the patch. The interpolation is governed by an adaptive weight function \( W^{i}(t, T^{i}_{\text{map}}(x)) \), which depends on the diffusion timestep \( t \) and the local transmission values \( T^{i}_{\text{map}}(x) \). This weight function is defined as:
\begin{equation}
\label{eq:weight_function_reorder}
W^{i}(t, T^{i}_{\text{map}}(x)) = \cos\left( \frac{t}{T} \cdot \frac{\pi}{2} \right) \cdot \exp(-a \cdot t \cdot T^{i}_{\text{map}}(x)),
\end{equation}
where $T$ is the total number of diffusion steps and \( a \) is a hyperparameter controlling the transition rate. The cosine term facilitates a gradual shift from the hazy image to the clear image as timestep $t$ decreases during the reverse process, while the exponential term modulates this transition based on the local physical prior encoded in \( T^{i}_{\text{map}}(x) \). This formulation ensures that in regions with low transmission (i.e., dense haze), \( U_t^{i}(x) \) remains close to the hazy image \( I^{i}(x) \) during the initial stages of the reverse process (large \(t\)). This simplifies the task to a more stable reconstruction problem by grounding the model in the available, albeit degraded, information. In contrast, in regions with high transmission (clearer areas), \( U_t^{i}(x) \) transitions more rapidly toward the clean image \( J_0^{i}(x) \), accelerating convergence and providing valuable contextual cues for restoring adjacent, more heavily hazed regions.

Unlike standard DDPMs that define the forward process by adding noise directly to the clean image $J_0$, the PIST forward process corrupts the intermediate target \( U_t \) with Gaussian noise:
\begin{equation}
\label{eq:noisy_state_reorder}
J_t(x) = \sqrt{\gamma_t} U_t(x) + \sqrt{1 - \gamma_t} \epsilon, \quad \epsilon \sim \mathcal{N}(0, \mathbf{I}),
\end{equation}
where \( \gamma_t = \prod_{s=1}^{t} \alpha_s \) defines the noise schedule from standard DDPM formulations, with $\alpha_s$ being the per-step noise schedule parameter.

During the reverse process, a network \( f_\theta(I, J_t, t) \) is trained to predict the noise \( \epsilon \) added at the corresponding forward step. The denoised state at step \( t-1 \) is then recovered from the noisy state \( J_t \) using the following update rule:
\begin{equation}
\label{eq:pist_reverse_update}
J_{t-1} = \frac{1}{\sqrt{\alpha_{t}}} \left( J_{t} - \frac{1 - \alpha_{t}}{\sqrt{1 - \gamma_{t}}} f_\theta(I, J_{t}, \gamma_{t}) \right) + \sqrt{1 - \alpha_{t}} \bar{\epsilon}_{t},
\end{equation}
where \( \bar{\epsilon}_{t} \sim \mathcal{N}(0, \mathbf{I}) \) is a sample of Gaussian noise added to maintain stochasticity. The denoising network $f_\theta$ is optimized with the following L1 loss objective:
\begin{equation}
\label{eq:pist_loss_reorder}
\mathcal{L}_{\text{PIST}} = \mathbb{E}_{I, J_0, \epsilon \sim \mathcal{N}(0, \mathbf{I}), t} \left\| f_\theta(I, J_t, t) - \epsilon \right\|_1.
\end{equation}
By targeting the physics-guided intermediate state \( U_t \) instead of the final clean image \( J_0 \), the network implicitly learns a denoising trajectory informed by the physical properties of haze. This leads to more efficient and higher-quality restoration, particularly in dense haze scenarios.

A critical component of PIST is the accurate estimation of the transmission map \( T_{\text{map}} \), which encodes the local haze density. To this end, our model, RPD-Diff, employs an enhanced Dark Channel Prior (DCP)~\cite{2011Single} estimation strategy. This strategy integrates edge-preserving filtering and sky-region handling to improve robustness, as detailed in Algorithm~\ref{alg:improved_dcp}. The resulting high-fidelity transmission map is crucial for the effectiveness of the PIST framework.

% --- CORRECTED ALGORITHM CODE ---
\begin{algorithm}[tbh]
\small
\caption{\small Improved Transmission Map Estimation with Sky Preservation}
\label{alg:improved_dcp}
% Note: No need for \baselinestretch or \itemsep here, 
% the algorithmic environment handles spacing.
\begin{algorithmic}[1] % The [1] enables line numbering
    \Require Hazy image \( I \), atmospheric light \( A \), parameters \( \omega, r, \epsilon, T_0, \tau_g, \tau_b \)
    \Ensure Transmission map \( T_{\text{map}} \)
    
    \State Compute dark channel: \( I_{\text{dark}}(x) \gets \min_{y \in \Omega(x)} \left( \min_{c \in \{r,g,b\}} I^c(y) \right) \)
    \State Compute initial transmission: \( \tilde{T}(x) \gets 1 - \omega \cdot \frac{I_{\text{dark}}(x)}{A} \)
    \State Convert \( I \) to grayscale image \( I_{\text{gray}} \) (using weighted averaging or PCA)
    \State Compute gradient map: \( G \gets \text{EdgeDetection}(I_{\text{gray}}) \)
    \State Denoise and filter \( G \) (using Gaussian or bilateral filtering)
    
    \State Generate sky mask \( S \):
    \If{\( G(x) < \tau_g \) \textbf{and} \( I_{\text{gray}}(x) > \tau_b \)}
        \State \( S(x) \gets 1 \)
    \Else
        \State \( S(x) \gets 0 \)
    \EndIf
    
    \State Apply Gaussian feathering to \( S \), yielding \( S_{\text{smooth}} \)
    \State Refine with Guided Filter: \( T_{\text{map}} \gets \text{GuidedFilter}(\tilde{T}, I_{\text{gray}}, r, \epsilon) \)
    \State Adjust sky region: \( T_{\text{map}}(x) \gets \frac{S_{\text{smooth}}(x) \cdot \tilde{T}(x) + T_{\text{map}}(x) \cdot (255 - S_{\text{smooth}}(x))}{255} \)
    \State Apply lower bound: \( T_{\text{map}}(x) \gets \max(T_{\text{map}}(x), T_0) \)
    
    \State \Return Transmission map \( T_{\text{map}}\)
\end{algorithmic}
\end{algorithm}

\subsection{Haze-Aware Denoising Timestep Predictor}
Non-uniform haze distribution induces spatial heterogeneity in degradation, implying that different image regions necessitate varying levels of denoising effort. Conventional DDPM frameworks typically apply a uniform timestep \( t \) throughout the image during the reverse process, which can hinder optimal restoration in regions with significantly different haze density. To overcome this limitation, we propose the Haze-Aware Denoising Timestep Predictor (HADTP), which dynamically modulates the denoising timestep and corresponding intensities for each image patch based on local haze characteristics.

As depicted in Fig.~\ref{fig:pipeline} (b), HADTP leverages a Transmission Map Cross-Attention (TMCA) mechanism to estimate a patch-specific timestep offset \( \Delta t \) relative to the current global timestep \( t \). The transmission map \( T_{\text{map}} \) serves as a spatially adaptive confidence map, which guides both the enhancement of conditional inputs and the adjustment of timesteps for individual patches. The TMCA module operates by applying a logically inverse strategy to the intermediate denoising result \( J_t^{(i)} \) and the input hazy image \( I^{(i)} \): (1) High-transmission regions in \textit{the intermediate denoising result} are enhanced, as these areas retain more accurate structural and color information. (2) Low-transmission regions in \textit{the hazy image} are prioritized for increased attention, directing greater denoising effort to these challenging areas.

The enhanced conditional input, derived from the TMCA mechanism, is subsequently concatenated with the intermediate denoising result \( J_t^{(i)} \) and the original hazy patch \( I^{(i)} \), which is then fed into the denoising network \( f_\theta \) for the subsequent iteration of the reverse diffusion process. The timestep offset \( \Delta t \) generated by HADTP adjusts the current timestep for each patch to \( \hat{t} = t + \Delta t \), enabling adaptive modification of the denoising intensity across the image. By conditioning the process on the estimated physical prior \( T_{\text{map}} \) and the current restoration state \( J_t^{(i)} \), HADTP effectively customizes the denoising schedule to account for local degradation patterns. This adaptability facilitates precise and efficient management of non-uniform haze distributions, enhancing the overall restoration quality in complex real-world scenarios.

\section{EXPERIMENTS}
\subsection{Experimental Setup}
\textbf{Dataset} \quad We conduct our experiments on four widely-used, real-world image dehazing benchmarks: Dense-Haze~\cite{densehaze}, NH-Haze~\cite{nhaze}, I-Haze~\cite{ihaze}, and O-Haze~\cite{ohaze}. Collectively, these datasets encompass a diverse range of challenging scenarios, featuring both dense and non-uniform haze distributions across indoor and outdoor environments. For each dataset, we designate the final five images for the test set and utilize the remaining images for training.

\textbf{Implementation Details} \quad We employ ADAM as the optimizer with \( \beta_1 = 0.9 \), \( \beta_2 = 0.999 \), and set the initial learning rate to \( 2 \times 10^{-4} \). The patch size is configured to \( p = 64 \), with a sliding stride of \( r = 16 \). The maximum number of denoising timestep \( T \) is configured to $1000$, with the Haze-Aware Denoising Timestep Predictor (HADTP) dynamically adjusting this timestep for each patch during inference. The denoising network adopts a U-Net architecture similar to \cite{stable}.

\textbf{Compared Models} \quad We compare our method quantitatively and qualitatively with several state-of-the-art (SOTA) algorithms, including one prior-based algorithm (DCP \cite{2011Single}), and six deep learning-based methods: DehazeDDPM \cite{yu2024highqualityimagedehazingdiffusion}, FFA-Net \cite{FFA}, MSBDN \cite{MSBDN}, Dehamer \cite{dehamer}, FCB \cite{WANG2024106281}, and DehazeFormer \cite{dehazeformer}. To ensure fairness, we obtain results by using the author-provided implementations in a consistent training-test configuration. The performance of these models is assessed using two distortion-based metrics, PSNR and SSIM \cite{psnrssim}, as well as two perceptual-based metrics, LPIPS \cite{LPIPS} and FID \cite{FID}.

\subsection{Experimental Results}
\subsubsection{Quantitative Evaluation}
Table~\ref{tab:t1} presents the quantitative results of various methods across four benchmark datasets. Our proposed RPD-Diff consistently outperforms competing approaches in both distortion-based and perceptual metrics. Notably, RPD-Diff exhibits a significant performance advantage in challenging dense and non-uniform haze scenarios. Specifically, in terms of PSNR, RPD-Diff surpasses the second-best method by 3.02 dB on the Dense-HAZE dataset \cite{densehaze} and by 2.06 dB on the NH-HAZE dataset \cite{nhaze}. Regarding the perceptual metric FID, our method also achieves the best performance across all datasets. This superior performance can be primarily attributed to the integration of physics-guided intermediate state targeting and the adaptive handling of spatial heterogeneity, which effectively address the challenges of severe information degradation and non-uniform haze distributions.

\begin{table*}[t]
\centering
\caption{Quantitative Results. The performance of models is evaluated using distortion-based metrics (PSNR, SSIM \cite{psnrssim}, higher values indicate better dehazing) and perception-based metrics (LPIPS \cite{LPIPS}, FID \cite{FID}, lower values indicate better quality). Optimal and suboptimal values are highlighted in \textbf{bold} and \underline{underlined}, respectively.}
\label{tab:t1}
% 使用 resizebox 强制缩放到文本宽度
\resizebox{\textwidth}{!}{%
\begin{tabular}{c|cccc|cccc|cccc|cccc}
\toprule[1.2pt]
Datasets & \multicolumn{4}{c|}{I-HAZE} & \multicolumn{4}{c|}{O-HAZE} & \multicolumn{4}{c|}{Dense-HAZE} & \multicolumn{4}{c}{NH-HAZE} \\
\cmidrule(lr){2-5} \cmidrule(lr){6-9} \cmidrule(lr){10-13} \cmidrule(lr){14-17}
Indicators & PSNR $\uparrow$ & SSIM $\uparrow$ & LPIPS $\downarrow$ & FID $\downarrow$ & PSNR $\uparrow$ & SSIM $\uparrow$ & LPIPS $\downarrow$ & FID $\downarrow$ & PSNR $\uparrow$ & SSIM $\uparrow$ & LPIPS $\downarrow$ & FID $\downarrow$ & PSNR $\uparrow$ & SSIM $\uparrow$ & LPIPS $\downarrow$ & FID $\downarrow$ \\
\midrule
DCP        & 16.04 & 0.72 & 0.33 & 237.28 & 14.62 & 0.68 & 0.35 & 196.41 & 11.75 & 0.42 & 0.60 & 343.65 & 12.49 & 0.51 & 0.42 & 128.90 \\
DehazeDDPM & 16.52 & 0.70 & 0.17 & 151.44 & 17.02 & 0.70 & 0.17 & 133.44 & \underline{17.04} & \underline{0.59} & 0.32 & \underline{171.06} & 20.38 & \underline{0.74} & 0.17 & \underline{121.06} \\
FFA-Net    & 16.89 & 0.74 & 0.22 & 171.01 & 17.04 & 0.59 & 0.26 & 171.06 & 11.07 & 0.40 & 0.51 & 413.22 & 12.33 & 0.48 & 0.33 & 374.31 \\
MSBDN      & 16.99 & 0.75 & 0.19 & 135.01 & 16.56 & 0.65 & 0.27 & 176.99 & 11.14 & 0.38 & 0.48 & 335.03 & 12.58 & 0.48 & 0.32 & 287.81 \\
Dehamer    & 17.02 & 0.64 & 0.18 & 163.94 & \underline{18.42} & \underline{0.72} & 0.17 & \underline{107.37} & 16.62 & 0.56 & \underline{0.28} & 223.65 & 20.66 & 0.70 & 0.18 & 138.49 \\
FCB        & \underline{17.35} & \underline{0.75} & \underline{0.14} & \underline{118.95} & 17.39 & 0.64 & \underline{0.16} & 137.70 & 13.16 & 0.49 & 0.35 & 298.42 & \underline{21.17} & 0.72 & \underline{0.17} & 133.60 \\
DehazeFormer & 16.83 & 0.65 & 0.19 & 127.09 & 17.30 & 0.67 & 0.23 & 127.28 & 13.09 & 0.27 & 0.38 & 378.74 & 12.70 & 0.31 & 0.23 & 313.77 \\
Ours       & \textbf{19.02} & \textbf{0.75} & \textbf{0.12} & \textbf{100.56} & \textbf{20.54} & \textbf{0.73} & \textbf{0.14} & \textbf{104.05} & \textbf{20.06} & \textbf{0.63} & \textbf{0.25} & \textbf{126.30} & \textbf{23.44} & \textbf{0.82} & \textbf{0.12} & \textbf{67.68} \\
\bottomrule[1.2pt]
\end{tabular}%
} % 结束 resizebox
\end{table*}

\begin{figure*}[t]
\vspace{-0.1cm}
\centering
\setlength{\abovecaptionskip}{0.0cm}
\includegraphics[width=\textwidth]{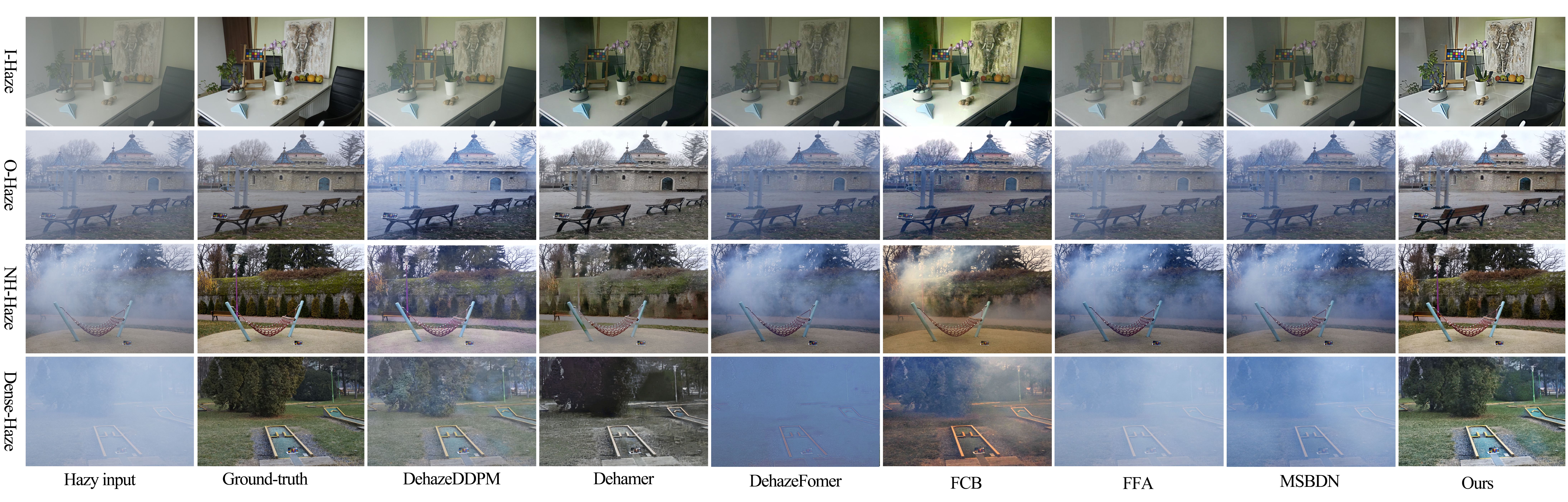}
\vspace{-1cm}
\caption{Qualitative results on the Dense-Haze \cite{densehaze}, Non-Homogeneous Haze \cite{nhaze}, O-Haze \cite{ohaze}, and I-Haze \cite{ihaze} datasets demonstrate the effectiveness of our approach. Under real-world conditions with dense and non-homogeneous haze, prior dehazing methods often result in residual haze and color distortions. In contrast, our RPD-Diff model produces high-quality, haze-free images with sharp edges and minimal haze artifacts.}
\label{fig:results}
\vspace{-1em}
\end{figure*}

\subsubsection{Qualitative Evaluation}
Fig.~\ref{fig:results} illustrates the dehazing performance of RPD-Diff compared to other SOTA methods across various real-world scenarios. In dense and non-uniform haze conditions, current methods often produce results with significant residual haze, noticeable artifacts, and pronounced color distortions. In contrast, RPD-Diff delivers visually superior results, characterized by enhanced detail clarity and improved color fidelity, demonstrating its robustness in handling complex haze distributions.

\subsubsection{Model Efficiency Analysis}
We evaluate the average computational efficiency of RPD-Diff against two SOTA diffusion-based dehazing methods across four datasets, as presented in Table~\ref{tab:eff}. RPD-Diff achieves the lowest inference time and FLOPs, owing to its patch-based generation approach, which effectively reduces computational complexity by dividing the large image into smaller patches, thereby minimizing the overall generation workload.

\subsubsection{Ablation Study}
We perform ablation studies across four datasets to investigate the contribution of each component. The results (Table \ref{tab:ablation}) demonstrate that both PIST and HADTP significantly improve image restoration quality with their combination achieving the optimal results. PIST improves restoration quality by reformulating the diffusion process with physics-guided intermediate targets, while HADTP enhances adaptability to non-uniform haze through dynamic timestep adjustments. Their integration results in a robust and effective dehazing framework, as demonstrated by the substantial performance gains across all evaluated metrics.

\begin{table}[h]
\setlength{\abovecaptionskip}{0cm}
\setlength{\belowcaptionskip}{-0.0cm}
\centering
\begin{minipage}{0.34\textwidth}
\caption{Efficiency Analysis}
\label{tab:eff}
\resizebox{\textwidth}{!}{
\begin{threeparttable}    
\begin{tabular}{*{4}{c}}
\toprule
Metrics & DehazeDDPM & FCB & Ours \\
\midrule
GPU Time $\downarrow$ & 16.4s & 8.5s & \textbf{3.2s} \\
FLOPs $\downarrow$ & 1102G & 614G & \textbf{325G} \\
\bottomrule
\end{tabular}
\begin{tablenotes}
\scriptsize
\item Note: All images are resized to a uniform dimension of $400\times600$ and processed by four NVIDIA 4090 GPUs.
\end{tablenotes}
\end{threeparttable}
}
\end{minipage}%
\hfill
\begin{minipage}{0.66\textwidth}
\caption{Ablation Experiments}
\label{tab:ablation}
\resizebox{\textwidth}{!}{
\begin{tabular}{*{6}{c}}
\toprule
PIST & HADTP & PSNR & SSIM & LPIPS & FID \\
\midrule
\ding{55} & \ding{55} & 19.05 & 0.74 & 0.15 & 84.67 \\
\ding{51} & \ding{55} & 20.34 $\textcolor{red}{(6.77\%\uparrow)}$ & 0.76 $\textcolor{red}{(2.72\%\uparrow)}$ & 0.14 $\textcolor{red}{(6.67\%\downarrow)}$ & 77.46 $\textcolor{red}{(8.52\%\downarrow)}$ \\
\ding{55} & \ding{51} & 21.89 $\textcolor{red}{(14.91\%\uparrow)}$ & 0.77 $\textcolor{red}{(4.05\%\uparrow)}$ & 0.12 $\textcolor{red}{(20.00\%\downarrow)}$ & 72.12 $\textcolor{red}{(14.82\%\downarrow)}$ \\
\ding{51} & \ding{51} & 23.44 $\textcolor{red}{(23.04\%\uparrow)}$ & 0.82 $\textcolor{red}{(10.81\%\uparrow)}$ & 0.12 $\textcolor{red}{(20.00\%\downarrow)}$ & 67.68 $\textcolor{red}{(20.07\%\downarrow)}$ \\
\bottomrule
\end{tabular}
}
\end{minipage}
\end{table}

\section{CONCLUSION}
In this work, we propose RPD-Diff, a Region-adaptive Physics-guided Dehazing Diffusion Model designed to address the challenges of visibility enhancement under dense and non-uniform haze conditions. By introducing the Physics-guided Intermediate State Targeting (PIST) strategy, RPD-Diff reformulates the diffusion process to mitigate the issue of insufficient conditioning in dense haze scenarios, enabling high-fidelity restoration. Furthermore, the Haze-Aware Denoising Timestep Predictor (HADTP) dynamically adapts the denoising timestep to spatially varying haze distributions, ensuring precise handling of non-uniform haze. Experimental results on four real-world datasets demonstrate that RPD-Diff achieves superior performance over SOTA methods with significant improvements in both distortion-based metrics and perceptual quality, which highlights the potential of RPD-Diff for robust dehazing in challenging real-world scenarios.

%-------------------------------------------------------------------------
% Bibliography
\bibliography{egbib}

\begin{thebibliography}{10}\itemsep=-1pt

\bibitem{ihaze}
C.~Ancuti, C.~O. Ancuti, R.~Timofte, and C.~De~Vleeschouwer.
\newblock I-haze: A dehazing benchmark with real hazy and haze-free indoor images.
\newblock In {\em Proceedings of the 19th International Conference on Advanced Concepts for Intelligent Vision Systems (ACIVS)}, pages 620--631, September 2018.

\bibitem{densehaze}
C.~O. Ancuti, C.~Ancuti, M.~Sbert, and R.~Timofte.
\newblock Dense-haze: A benchmark for image dehazing with dense-haze and haze-free images.
\newblock In {\em Proceedings of the IEEE International Conference on Image Processing (ICIP)}, pages 1014--1018, September 2019.

\bibitem{nhaze}
C.~O. Ancuti, C.~Ancuti, and R.~Timofte.
\newblock Nh-haze: An image dehazing benchmark with non-homogeneous hazy and haze-free images.
\newblock In {\em Proceedings of the IEEE/CVF Conference on Computer Vision and Pattern Recognition Workshops (CVPRW)}, pages 1798--1805, June 2020.

\bibitem{ohaze}
C.~O. Ancuti, C.~Ancuti, R.~Timofte, and C.~De~Vleeschouwer.
\newblock O-haze: A dehazing benchmark with real hazy and haze-free outdoor images.
\newblock In {\em Proceedings of the IEEE/CVF Conference on Computer Vision and Pattern Recognition Workshops (CVPRW)}, pages 867--8678, June 2018.

\bibitem{Cai_Xu_Jia_Qing_Tao_2016}
B.~Cai, X.~Xu, K.~Jia, C.~Qing, and D.~Tao.
\newblock Dehazenet: An end-to-end system for single image haze removal.
\newblock {\em IEEE Transactions on Image Processing}, 25(11):5187--5198, November 2016.

\bibitem{asm}
A.~Cantor.
\newblock Optics of the atmosphere--scattering by molecules and particles.
\newblock {\em IEEE Journal of Quantum Electronics}, 14(9):698--699, 1978.

\bibitem{2016Robust}
C.~Chen, M.~N. Do, and J.~Wang.
\newblock Robust image and video dehazing with visual artifact suppression via gradient residual minimization.
\newblock In B.~Leibe, J.~Matas, N.~Sebe, and M.~Welling, editors, {\em Computer Vision -- ECCV 2016}, pages 576--591, Cham, 2016. Springer International Publishing.

\bibitem{NEURIPS2021_49ad23d1}
P.~Dhariwal and A.~Q. Nichol.
\newblock Diffusion models beat gans on image synthesis.
\newblock In {\em Proceedings of the 35th International Conference on Neural Information Processing Systems (NeurIPS)}, pages 8780--8794, December 2021.

\bibitem{Raanan2014Dehazing}
R.~Fattal.
\newblock Dehazing using color-lines.
\newblock {\em ACM Transactions on Graphics}, 34(1), December 2015.

\bibitem{GDP}
B.~Fei, Z.~Lyu, L.~Pan, J.~Zhang, W.~Yang, T.~Luo, B.~Zhang, and B.~Dai.
\newblock Generative diffusion prior for unified image restoration and enhancement.
\newblock In {\em Proceedings of the IEEE/CVF Conference on Computer Vision and Pattern Recognition (CVPR)}, pages 9935--9946, June 2023.

\bibitem{dehamer}
C.~Guo, Q.~Yan, S.~Anwar, R.~Cong, W.~Ren, and C.~Li.
\newblock Image dehazing transformer with transmission-aware 3d position embedding.
\newblock In {\em Proceedings of the IEEE/CVF Conference on Computer Vision and Pattern Recognition (CVPR)}, pages 5802--5810, 2022.

\bibitem{2011Single}
K.~He, J.~Sun, and X.~Tang.
\newblock Single image haze removal using dark channel prior.
\newblock {\em IEEE Transactions on Pattern Analysis and Machine Intelligence}, 33(12):2341--2353, December 2011.

\bibitem{FID}
M.~Heusel, H.~Ramsauer, T.~Unterthiner, B.~Nessler, and S.~Hochreiter.
\newblock Gans trained by a two time-scale update rule converge to a local nash equilibrium.
\newblock In {\em Proceedings of the 31st International Conference on Neural Information Processing Systems (NeurIPS)}, pages 6629--6640, December 2017.

\bibitem{DDPM}
J.~Ho, A.~Jain, and P.~Abbeel.
\newblock Denoising diffusion probabilistic models.
\newblock {\em Advances in Neural Information Processing Systems}, 33:6840--6851, December 2020.

\bibitem{DC-Scence}
T.~{Huang}, Z.~{Zhang}, R.~{Zhang}, and Y.~{Zhao}.
\newblock {DC-Scene: Data-Centric Learning for 3D Scene Understanding}.
\newblock {\em arXiv e-prints}, page arXiv:2505.15232, May 2025.

\bibitem{2017AOD}
B.~Li, X.~Peng, Z.~Wang, J.~Xu, and D.~Feng.
\newblock Aod-net: All-in-one dehazing network.
\newblock In {\em Proceedings of the IEEE International Conference on Computer Vision (ICCV)}, pages 4780--4788, October 2017.

\bibitem{Liu_Ma_Shi_Chen_2019}
X.~Liu, Y.~Ma, Z.~Shi, and J.~Chen.
\newblock Griddehazenet: Attention-based multi-scale network for image dehazing.
\newblock In {\em Proceedings of the IEEE/CVF International Conference on Computer Vision (ICCV)}, pages 7313--7322, October 2019.

\bibitem{Liu_Pan_Ren_Su_2019}
Y.~Liu, J.~Pan, J.~Ren, and Z.~Su.
\newblock Learning deep priors for image dehazing.
\newblock In {\em Proceedings of the IEEE/CVF International Conference on Computer Vision (ICCV)}, pages 2492--2500, October 2019.

\bibitem{Lugmayr_2022_CVPR}
A.~Lugmayr, M.~Danelljan, A.~Romero, F.~Yu, R.~Timofte, and L.~Van~Gool.
\newblock Repaint: Inpainting using denoising diffusion probabilistic models.
\newblock In {\em Proceedings of the IEEE/CVF Conference on Computer Vision and Pattern Recognition (CVPR)}, pages 11461--11471, June 2022.

\bibitem{niqe}
A.~Mittal, R.~Soundararajan, and A.~C. Bovik.
\newblock Making a “completely blind” image quality analyzer.
\newblock {\em IEEE Signal Processing Letters}, 20(3):209--212, March 2013.

\bibitem{FFA}
X.~Qin, Z.~Wang, Y.~Bai, X.~Xie, and H.~Jia.
\newblock Ffa-net: Feature fusion attention network for single image dehazing.
\newblock In {\em Proceedings of the AAAI Conference on Artificial Intelligence}, volume~34, pages 11908--11915, April 2020.

\bibitem{Qin_Wang_Bai_Xie_Jia_2020}
X.~Qin, Z.~Wang, Y.~Bai, X.~Xie, and H.~Jia.
\newblock Ffa-net: Feature fusion attention network for single image dehazing.
\newblock {\em Proceedings of the AAAI Conference on Artificial Intelligence}, 34(7):11908--11915, April 2020.

\bibitem{jinzhi_ICCV}
Y.~Qiu, K.~Zhang, C.~Wang, W.~Luo, H.~Li, and Z.~Jin.
\newblock Mb-taylorformer: Multi-branch efficient transformer expanded by taylor formula for image dehazing.
\newblock In {\em Proceedings of the IEEE/CVF International Conference on Computer Vision (ICCV)}, pages 12756--12767, October 2023.

\bibitem{stable}
R.~Rombach, A.~Blattmann, D.~Lorenz, P.~Esser, and B.~Ommer.
\newblock High-resolution image synthesis with latent diffusion models.
\newblock In {\em Proceedings of the IEEE/CVF conference on computer vision and pattern recognition (CVPR)}, pages 10684--10695, 2022.

\bibitem{Saharia_Ho_Chan_Salimans_Fleet_Norouzi_2022}
C.~Saharia, J.~Ho, W.~Chan, T.~Salimans, D.~J. Fleet, and M.~Norouzi.
\newblock Image super-resolution via iterative refinement.
\newblock {\em IEEE Transactions on Pattern Analysis and Machine Intelligence}, 45(4):4713--4726, May 2023.

\bibitem{dehazeformer}
Y.~Song, Z.~He, H.~Qian, and X.~Du.
\newblock Vision transformers for single image dehazing.
\newblock {\em IEEE Transactions on Image Processing}, 32:1927--1941, April 2023.

\bibitem{MSBDN}
G.~Wang and X.~Yu.
\newblock Msf2dn: Multi scale feature fusion dehazing network with dense connection.
\newblock In {\em Proceedings of the Asian Conference on Computer Vision (ACCV)}, pages 444--459, December 2022.

\bibitem{WANG2024106281}
J.~Wang, S.~Wu, Z.~Yuan, Q.~Tong, and K.~Xu.
\newblock Frequency compensated diffusion model for real-scene dehazing.
\newblock {\em Neural Networks}, 175:106281, 2024.

\bibitem{psnrssim}
Z.~Wang, A.~C. Bovik, H.~R. Sheikh, and E.~P. Simoncelli.
\newblock Image quality assessment: from error visibility to structural similarity.
\newblock {\em IEEE Transactions on Image Processing}, 13(4):600--612, April 2004.

\bibitem{ssim}
Z.~Wang, A.~C. Bovik, H.~R. Sheikh, and E.~P. Simoncelli.
\newblock Image quality assessment: From error visibility to structural similarity.
\newblock {\em IEEE Transactions on Image Processing}, 13(4):600--612, April 2004.

\bibitem{DriftRec}
S.~Welker, H.~N. Chapman, and T.~Gerkmann.
\newblock Driftrec: Adapting diffusion models to blind jpeg restoration.
\newblock {\em IEEE Transactions on Image Processing}, 33:1726--1738, January 2024.

\bibitem{yu2024highqualityimagedehazingdiffusion}
H.~Yu, J.~Huang, K.~Zheng, and F.~Zhao.
\newblock High-quality image dehazing with diffusion model.
\newblock {\em arXiv preprint arXiv:2308.11949}, August 2023.

\bibitem{FSDGN}
H.~Yu, N.~Zheng, M.~Zhou, J.~Huang, Z.~Xiao, and F.~Zhao.
\newblock Frequency and spatial dual guidance for image dehazing.
\newblock In S.~Avidan, G.~Brostow, M.~Ciss{\'e}, G.~M. Farinella, and T.~Hassner, editors, {\em Computer Vision -- ECCV 2022}, pages 181--198, Cham, 2022. Springer Nature Switzerland.

\bibitem{2017Fast}
J.~Zhang, Y.~Cao, S.~Fang, Y.~Kang, and C.~W. Chen.
\newblock Fast haze removal for nighttime image using maximum reflectance prior.
\newblock In {\em Proceedings of the IEEE Conference on Computer Vision and Pattern Recognition (CVPR)}, pages 7016--7024, July 2017.

\bibitem{DAPM}
L.~Zhang, W.~Bai, and C.~Xiao.
\newblock Density-aware diffusion model for efficient image dehazing.
\newblock {\em Computer Graphics Forum}, 43(7):e15221, October 2024.

\bibitem{LPIPS}
R.~Zhang, P.~Isola, A.~A. Efros, E.~Shechtman, and O.~Wang.
\newblock The unreasonable effectiveness of deep features as a perceptual metric.
\newblock In {\em Proceedings of the IEEE Conference on Computer Vision and Pattern Recognition (CVPR)}, pages 586--595, June 2018.

\bibitem{FDG-Diff}
R.~Zhang, K.~Tian, Z.~Zhang, Q.~Liu, and Z.~Jin.
\newblock Fdg-diff: Frequency-domain-guided diffusion framework for compressed hazy image restoration.
\newblock {\em arXiv preprint arXiv:2501.12832}, January 2025.

\bibitem{T_fusion}
S.~Zhao, X.~Zhang, P.~Xiao, and G.~He.
\newblock Exchanging dual encoder-decoder: A new strategy for change detection with semantic guidance and spatial localization.
\newblock {\em IEEE Transactions on Geoscience and Remote Sensing}, 61:1--16, November 2023.

\bibitem{zhou2025fireedit}
J.~Zhou, J.~Li, Z.~Xu, H.~Li, Y.~Cheng, F.-T. Hong, Q.~Lin, Q.~Lu, and X.~Liang.
\newblock Fireedit: Fine-grained instruction-based image editing via region-aware vision language model.
\newblock {\em arXiv preprint arXiv:2503.19839}, March 2025.

\bibitem{patchDDPM}
O.~Özdenizci and R.~Legenstein.
\newblock Restoring vision in adverse weather conditions with patch-based denoising diffusion models.
\newblock {\em IEEE Transactions on Pattern Analysis and Machine Intelligence}, 45(8):10346--10357, August 2023.

\end{thebibliography}

\end{document}